\documentclass[journal]{IEEEtran}

\usepackage{cite}

\ifCLASSINFOpdf
\else
\fi
\usepackage{graphicx}

\usepackage{amsmath}
\usepackage{amssymb}          
\DeclareMathOperator*{\argmax}{argmax}

\usepackage{array}
\usepackage{hyperref} 
\hypersetup{
    colorlinks=true,
    linkcolor=black,
    filecolor=black, 
    citecolor=blue,
    urlcolor=blue,
}

\begin{document}
%
\title{Behavior coordination for self-adaptive robots using constraint-based configuration}

%
%
%

\author{Martin~Molina, Pablo~Santamaria 
\thanks{Authors affiliation: Department of Artificial Intelligence,  \textit{Universidad Politécnica de Madrid}, Spain. Research Group CVAR (Computer Vision and Aerial Robotics).}
}

\maketitle

\thispagestyle{empty}

\begin{abstract}
Autonomous robots may be able to adapt their behavior in response to changes in the environment. This is useful, for example, to efficiently handle limited resources or to respond appropriately to unexpected events such as faults. The architecture of a self-adaptive robot is complex because it should include automatic mechanisms to dynamically configure the elements that control robot behaviors. To facilitate the construction of this type of architectures, it is useful to have general solutions in the form of software tools that may be  
applicable to different robotic systems. This paper presents an original algorithm to dynamically configure the control architecture, which is applicable to the development of self-adaptive autonomous robots. This algorithm uses a constraint-based configuration approach to decide which basic robot behaviors should be activated in response to both reactive and deliberative events. The algorithm uses specific search heuristics and initialization procedures to achieve the performance required by robotic systems. The solution has been implemented as a software development tool called Behavior Coordinator CBC (Constraint-Based Configuration), which is based on ROS and open source, available to the general public. This tool has been successfully used for building multiple applications of autonomous aerial robots.

\end{abstract}


%

\section{Introduction}
%
%
%
%

\begin{figure}[htb!]
\begin{center}
\includegraphics[width=0.50\textwidth]{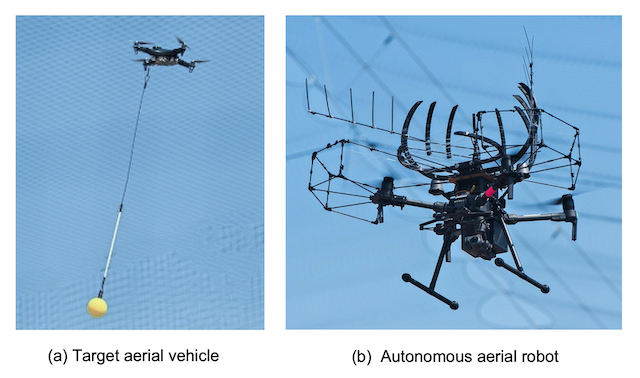}
\caption{An autonomous robot exhibits self-adaptive behavior in a visual-based target following mission to efficiently use limited resources \cite{Suarez2020}. This figure shows (a) the target aerial vehicle carrying a ball, and (b) the autonomous robot to catch the ball.}
\label{fig:drones mbzirc}
\end{center}
\end{figure}

\IEEEPARstart{S}{elf-adaptive} systems are able to modify their behavior in response to changes in the environment. Figure \ref{fig:drones mbzirc} shows an example of self-adaptive autonomous robot that performs a vision-based target following mission\footnote{This autonomous robot was developed in our Research Group CVAR \cite{Suarez2020} to participate in the International Competition MBZIRC 2020.}. In this mission, the goal of the robot is to catch a ball carried by another aerial vehicle (the figure shows the autonomous robot and the aerial vehicle with the ball). 

This system exhibits self-adaptive behavior in two aspects. First, when the robot is flying far away from the target vehicle, it moves with slow accelerations to save energy. However, when the robot is close to the target, it performs aggressive maneuvers with high speeds and accelerations in order to be able to intercept the target. Second, this robot uses two different visual detection algorithms to recognize the target vehicle (calibrated to different distances to the target). Since both detection algorithms cannot be executed at the same time due to their high computational load, the robot changes the visual recognition method according to the distance to the target.

This example shows a basic form of self-adaptation to efficiently manage the robot's limited resources. However, self-adaptive capability can improve autonomous robot operation in other ways. For example, self-adaptation can improve reliability by automatically modifying the system’s behavior in response to detected faults. Besides, self-adaptive robots may be easier to use, because humans operators can specify robot’s goals using more abstract tasks whose execution is automatically adapted to the conditions of the environment during the mission execution. 

However, the advantages of self-adaptive systems come at the price of a more complex control architecture, whose development creates additional technical difficulties. Self-adaptation can be achieved in robotics with dynamic architectures with sets of alternative methods to control robot behaviors. In the previous example, the aerial robot includes two alternative methods for motion control and two alternative methods for visual recognition. The architecture should also include an adaptation logic to dynamically select the most appropriate methods for each case. 

This paper presents an original algorithm to dynamically configure the control architecture, which is applicable to the development of self-adaptive autonomous robots. In our work, self-adaptation is done by activating and deactivating basic robot behaviors. Our algorithm uses a constraint-based configuration approach to decide which behaviors should be activated in response to both reactive events (e.g., unexpected events caused by faults) and deliberative events (e.g., requested tasks).

The solution presented in this paper has been implemented in a software development tool called Behavior Coordinator CBC (Constraint-Based Configuration) that is based on ROS (Robot Operating System) and it is open source available for the general public. This tool has been conceived and developed as the evolution and generalization of previous work about tools for building autonomous aerial robots \cite{Molina2019}\cite{Molina2020}. 

The remainder of this paper is organized in the following way. Section 2 describes related work about software development tools for building self-adaptive systems. Section 3 describes an overview of our approach for self-adaptation. Section 4 describes the algorithm that we have designed for automatic self-configuration. Section 5 describes implementation details of the software development tool. Section 6 presents evaluation results about applicability, reusability and performance efficiency. Finally, section 7 presents general conclusions about this tool.

\section{Related work}
Self-adaptation have been studied for a long time in fields such as artificial intelligence and robotics. Some of the first solutions for self-adaptation were proposed more than three decades ago. For example, Maes proposed a solution for adaptation that selects dynamically executable competence modules \cite{Maes1989}. Firby presented the idea of reactive action packages (RAPs) as a solution for adaptive execution in complex dynamic worlds \cite{Firby1989}. Ferguson proposed a general architecture for intelligent adaptive agents \cite{Ferguson1992}. Hayes-Roth presented an architecture for adaptive intelligent systems and identified different categories for adaptation \cite{HayesRoth1995}. Sroulia and Goel describe a reflective approach for self-adaptation using structure-behavior-function models to represent how the components work\cite{Stroulia1995}.

Self-adaptive systems have received important attention from the side of the research community in software systems. For example, Oreizy \textit{et al.} analyze the concept of self-adaptive software and specify general functions and components for system adaptation \cite{Oreizy1999}. General research challenges in self-adaptive software systems have been analyzed and identified by different works \cite{Cheng2009} \cite{Salehie2009} \cite{deLemos2013}. 

Krupitzer \textit{et al.} published a survey on engineering approaches for self-adaptive systems \cite{Krupitzer2015} with a taxonomy that identifies multiple categories in which self-adaptation methods can be classified. According to this taxonomy, our approach has an adaptation logic that is external, separated from the managed resources. The managed resources in our work are robot behaviors as they are considered in behavior-based robotics. The adaptation decision criteria uses a constraint-based representation to find consistent configurations of active behaviors. 

Other proposals for self-adaptive systems have also used a constraint-based representation for automatic configuration in self-adaptive systems. For example, Garlan \textit{et al.}\cite{Garlan2004} use a constraint evaluation module for self-adaptation in general software systems (e.g., a web-based client-server system or a videoconferencing system). However, instead of having a general constraint satisfaction algorithm, they use domain-specific adaptation strategies to solve constraint violations. Parra \textit{et al.} \cite{Parra2012} present a general constraint-based optimization method to support self-adaptation (applied to a problem of home control with mobile devices) with a representation based on dynamic software product lines.  Sawyer \textit{et al.} \cite{Sawyer2012} use constraint programming to propose a general approach for self-adaptive systems (applied to the management of a wireless sensor network). This approach is also based on the concept of dynamic software product lines. In comparison to these proposals, we use a different type of constraint-based representation with elements of multi-layered architectures of autonomous robots. Besides, our solution is implemented as a general software development tool that is freely available to the community of robotics.

There are ROS-based tools that can help building the architecture of self-adaptive robotic systems. The main difference between these tools and our proposal in the way they represent knowledge about the resources to manage. For example, MROS \cite{HernandeCorbato2020} is a comprehensive approach for self-adaptive robots that is presented as a ROS-based structured model-based framework for the adaptation of robot control architectures at run-time. The adaptation logic of this approach uses a model formulated using OWL (Ontology Web Language) extended with a rule-based representation and quality attributes \cite{HernandezCorbato2020metacontrol}. SkiROS \cite{Rovida2017} is a skill-based robot control platform on top of ROS in which skills that are selected dynamically using a planner based on PDDL. FlexBE \cite{Schillinger2016} can help building self-adaptive robots with monitoring capabilities and runtime control for safe execution. This is achieved with a representation based on state machines with additional mechanisms (e.g., locking procedures and consistency requirements). DyKnow \cite{DeLeng2016} is an extension to ROS that is able to reconfigure dynamically how components are connected to each other (changing dynamically) the flow of information in the system.

\section{General approach}
\begin{figure*}[htb!]
\begin{center}
\includegraphics[width=0.75\textwidth]{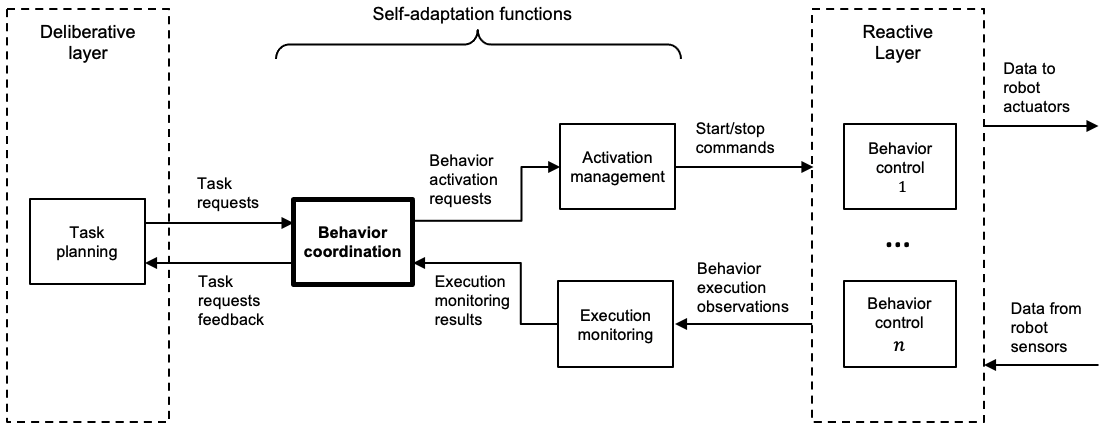}
\caption{Self-adaptation can be performed with three functions  (behavior coordination, activation management and execution monitoring) between a reactive layer and a deliberative layer. The constraint satisfaction algorithm presented in this paper corresponds to behavior coordination.}
\label{fig:self activation functions}
\end{center}
\end{figure*}

The approach followed in this paper for self-adaptation is based on the hybrid deliberative-reactive paradigm of robotics. According to this paradigm, the control architecture of an autonomous robot includes a reactive layer with a set of sensory-motor units called \textit{behaviors }\cite{Brooks1986} \cite{Arkin1998}. Each behavior operates in a perception-action loop and can be activated or deactivated individually during the execution of a mission\footnote{In our work, behaviors correspond to a generalization of the conventional concept of reactive behaviors (e.g., for motion control) in order to consider other kind of robot skills (e.g., visual recognition, simultaneous localization and mapping, etc.).}. Besides the reactive layer, the architecture has a deliberative layer, which includes reasoning methods (e.g., automated planning) that generate tasks to be done in order to achieve general goals. 

The deliberative and reactive layers can be integrated using an intermediate component called executive \cite{Kortenkamp2008}. The presence of this component is a solution for self-adaptation because the executive is able to adapt the global robot behavior by activating and deactivating behaviors (at the reactive layer) as a response of requested tasks (at the deliberative layer) taking into account changes in the dynamic environment (e.g., such as unexpected events). In our work, the term \textit{behavior coordination} \cite{Pirjanian1999} designates the decision of which behaviors should be activated in each situation\footnote{In contrast to other work, our approach for behavior coordination does not include command fusion \cite{Saffiotti1997}.}.

The algorithm presented in this paper considers behavior coordination a constraint-based configuration problem \cite{Junker2006}. In general, in a configuration problem, a set of components are combined and customized to meet certain requirements. In our case, the components are behaviors and the requirements are established by the requested tasks and the robot environment. 

According to our view, behavior coordination contributes to self-adaptation with other two additional functions (Figure \ref{fig:self activation functions}): (1) \textit{activation management}, which handles how to activate and deactivate robot behaviors, and (2) \textit{execution monitoring}, which supervises the execution of behavior controllers considering the state of the environment. 

Execution monitoring checks, for example, whether the current state of the environment satisfies the assumptions of behaviors to operate correctly (for example, in the case of a robot behavior that performs visual recognition, the monitoring function can check that the lighting assumptions in the environment are verified). Besides, execution monitoring detects when a behavior has reached the goal or the presence of unexpected events. For example, an abnormal operation can be detected if the robot does not arrive to a destination point in a maximum expected time or if the robot moves in the opposite direction to the direction that is expected to follow.

The division of self-adaptation in these three functions (behavior coordination, execution monitoring and activation management) is consistent with other approaches for self-adaptation. For example, the autonomic computing cycle for self-adaptive software systems \cite{Kephart2003} identifies four activities in a loop: monitor, analyze, plan, and execute (\textit{MAPE}). In our approach, execution monitoring corresponds to the activities \textit{M} (monitor) and \textit{A} (analyze), behavior coordination corresponds to \textit{P} (plan), and activation management corresponds to \textit{E} (execute).

\section{Concepts used in behavior coordination}
\begin{figure}[htb!]
\begin{center}
\includegraphics[width=0.40\textwidth]{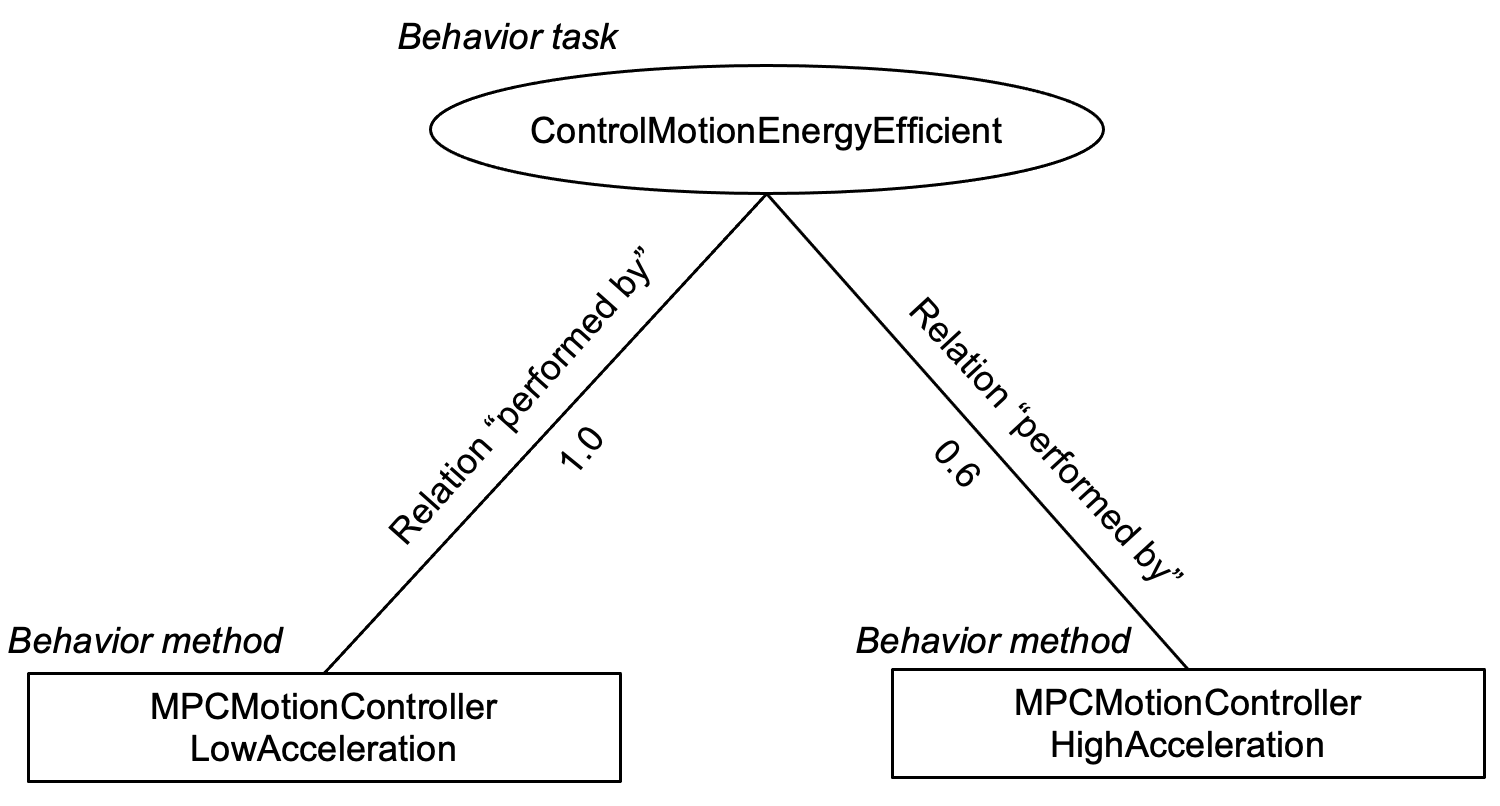}
\caption{Separation between task and method in a robot behavior.}
\label{fig:task method division}
\end{center}
\end{figure}

In this work, we use certain concepts about behavior coordination with a specific terminology. In a robot behavior, we distinguish the \textit{task} and the \textit{behavior method} (or \textit{behavior} in short). The task expresses \textit{what} the robot performs and the behavior method identifies \textit{how} the task is performed. Figure \ref{fig:task method division} shows an example of these concepts. The figure shows a task for control motion, called \(ControlMotionEnergyEfficient\), that can be performed by two different behaviors. The behavior called \(MotionControllerLowAcceleration\)  uses a motion controller with an algorithm based on MPC (model predictive control) that is configured with with a specific model to operate the robot with low accelerations. The other behavior also uses a MPC algorithm but, in this case, it is configured with a different model to provide high accelerations. 

In general, the name of the task describes a functionality to be done and, optionally, the name may also include non-functional requirements (e.g., requirements related to reliability or performance efficiency). In the example of Figure \ref{fig:task method division}, the task expresses that the motion control should be energy efficient. The first method with lower accelerations satisfies the requirements of this task in a better degree than the second. This is represented here with a quantitative value (1.0) for the first method. The second method has a lower value (0.6) because it applies a motion strategy with higher accelerations that require more energy.

Behavior coordination is performed in response to two different types of events. The first type of event corresponds to a request to start (or stop) a task. This event may be generated by an automated task planner or, in some cases, directly by a human operator that asks the robot to perform a certain task. The second type of event is generated by the execution monitoring function described above and corresponds to changes in the execution of behaviors. For example, this happens when a behavior finishes its execution with success (e.g., when the robot behavior has completed correctly a task) or failure (e.g., due to a change in the environment or an internal robot fault). 

\begin{figure}[htb!]
\begin{center}
\includegraphics[width=0.45\textwidth]{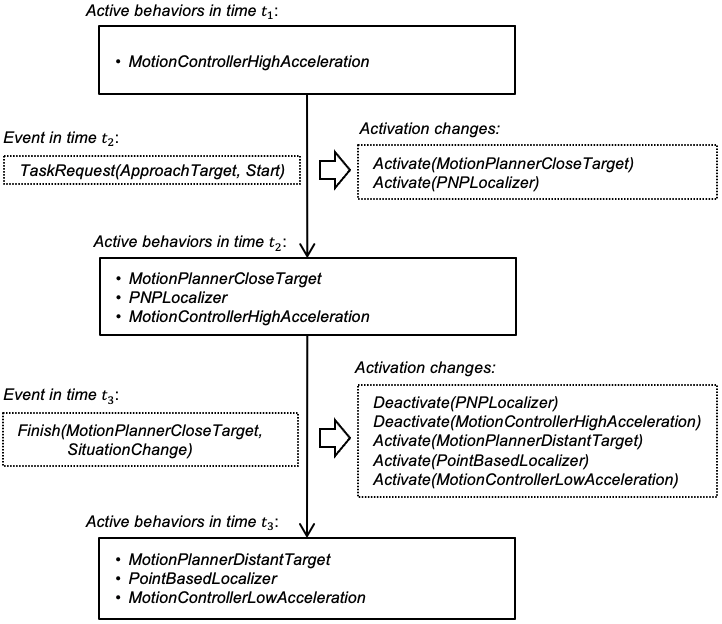}
\caption{Example of operation with a sequence of two events and the corresponding responses generated by behavior coordination.}
\label{fig:activations sequence}
\end{center}
\end{figure}

The result of behavior coordination is the set of behaviors that should be active in response to the events mentioned above (task requests or behavior execution changes), taking into account the relationships between the behaviors (e.g., about compatibility or dependency). This result can be expressed as a set of changes to be done in the current set of active behaviors. Figure \ref{fig:activations sequence} illustrates this operation using the visual-based target following mission. The figure shows a sequence of two events and the corresponding responses generated by behavior coordination. Initially, (in time \(t_1\)) there is only one active behavior, \(MotionControllerHighAcceleration\). Then, in time \(t_2\), a task is requested to start (task \(ApproachTarget\)). As a result of this request, behavior coordination generates two behavior activations (in this example, these two behaviors have been selected to perform the requested task because the target to follow is near the autonomous robot). Then, in time \(t_3\), an event is received corresponding to behavior \(MotionPlannerCloseTarget \) expressing that has finished its execution due to a  change in the situation (because the target to follow has been lost). As a response of this event, behavior coordination generates new changes in the set of active behaviors by deactivating two behaviors and activating other three behaviors.

\subsection{Starting methods of tasks}

Our approach for behavior coordination uses two task properties related to the way a task can be started. In general, a task may be started because it is requested by an external event (e.g., a task planner or by the human operator). In addition, a task could be started as a consequence of other reasons (e.g., because it is required by a behavior). For example, the task \(SelfLocalize\) can start because it is required by the activation of behavior \(RotatePIDControl\). It is important to distinguish the tasks that are not able to start without explicit request (e.g., \(FollowPath\) and \(TakeOff\)) because this property will be used by the constraint satisfaction algorithm to avoid unnecessary search (see explanation below). In our model, we use the property \textit{start on request} to distinguish tasks that can only be started on request.

A second task property is identified with the term \textit{reactive start}. A task with this property means that this task should be started whenever the conditions of the environment make it possible. This task is started if there is a behavior able to perform the task whose execution conditions are compatible with the current situation of the environment and, furthermore, there are no other tasks in execution that are incompatible with the task. For example, the task \(Hover\), used in aerial robotics, may have this property. The goal of this task is to maintain the aerial robot in a motionless flight over a reference point at a constant altitude. This task is automatically started whenever the robot is on the air and there are not other incompatible motion tasks started. It is assumed that a task that starts only by request cannot have the reactive start property, but there may be tasks that do not have any of these two properties.

\subsection{Priority scheme for task requests}

As it was mentioned above, a task request may be done from different sources such as an automated task planner or a human operator. To adequately coordinate multiple sources of task requests we use a priority level represented with integer numbers. In this scheme, lower numbers represent lower priority. For example, we use the following numbers in our applications for aerial robotics: 1 for a task planner, 2 for the human operator and 3 for a safety monitor. In the case of requests with the same level of priority, more recent requests have more priority.

\section{Constraint-based representation} \label{sec: constraint-based representation}
The algorithm that we present for behavior coordination is formulated as a constraint-based configuration problem \cite{Junker2006} using the representation described in this section.

This representation\footnote{We follow here the notation used by \cite{Rossi2006}.}  uses a set of variables \(X = \{x_1, x_2, ..., x_n\}\) where each variable \(x_i\) corresponds to a task. The set of possible values for each variable \(x_i\) is the domain \(D(x_i)  = \{ \varnothing, b_1, b_2, ..., b_k\}\), i.e., the set of all behaviors \( b_1, b_2, ..., b_k\) that can perform the task \(x_i\) with an additional element, expressed with the symbol \(\varnothing\), that represents the situation when the task is not started (i.e., when the task is not performed by any behavior). 

With this representation, it is assumed that each task \(x_i\) can potentially be performed by different behaviors and, at a given moment, task \(x_i\) is performed by only one active behavior.  A solution to the constraint satisfaction problem is expressed as an assignment of a single value  \(v_i \in D(x_i)\)  to each variable in the following way:
\begin{align*}
A = \{ x_1 = v_1, x_2 = v_2, ..., x_n = v_n\}
\end{align*}

\subsection{Compatibility and requirement constraints}

We have identified two main types of constraints in behavior coordination, that are usual in configuration problems \cite{Junker2006}. On the one hand, there are \textit{compatibility constraints} expressing that two tasks cannot be executed at the same time. For example, two motion tasks (e.g., taking off and landing) are incompatible because they cannot be performed at the same time. A compatibility constraint can be formulated as follows (meaning that at least one of two incompatible tasks \(x_i\) and \(x_j\) must not be running):
\begin{align*}
C_k:(x_i = \varnothing)\vee (x_j =\varnothing)
\end{align*}

On the other hand, there are \textit{requirement constraints}, which are used to express that the execution of a behavior requires to perform a certain task. For example, the execution of a method for motion control may require to perform a self-localization task. A requirement constraint can be formulated in the following way (meaning that behavior \(b_i\) that performs task \(x_i\) requires task \(x_j\)). 
\begin{align*}
C_k:(x_i=b_i)\rightarrow(x_j \neq \varnothing) 
\end{align*}

It is assumed that requirement constraints satisfy the following property derived from the characteristics of the behavior coordination problem. Let be the set \(R_i\) that represents the set of tasks that are required by task \(x_i\) (directly or indirectly through multiple requirement constraints) considering a particular assignment \(A\). For example, if we have the assignment \(A = \{ x_1 = b_1, x_2 =  \varnothing,  x_3 = b_3,  x_4 = b_4\}\) and two requirement constraints \(C_1:(x_1=b_1)\rightarrow(x_3\neq \varnothing) \)  and \(C_2:(x_3=b_3)\rightarrow(x_4\neq \varnothing) \), the set of required tasks for the task \(x_1\) is \(R_1 = \{x_3, x_4\}\). In general, it is assumed that \(x_i \notin R_i\) for all possible assignments. This means that the paths between tasks established by requirement constraints should not present loops. Instead, they organize tasks in a hierarchical structure where upper level tasks require lower level tasks. 

\subsection{Performance constraints}

Besides the previous constraints, there are other types of constraints related to the quality of a configuration. This quality is affected by the fact that the selected behaviors may satisfy tasks requirements in a partial degree. It may be needed to represent constraints in order to avoid potential behavior configurations that, as a whole, present low performance. 

To establish this type of constraints, we define a function \(\sigma(b _i, x_i) \in [0, 1]\), called suitability, that represents the degree in which behavior \(b _i\) satisfies the requirements of task \(x_i\). The value \(\sigma(b _i, x_i) = 1\) corresponds to the maximum degree of satisfaction. 

We can estimate the degree of performance of a task \(x_i\) considering not only the suitability of the behavior assigned to the task but also the suitability of the behaviors assigned to the tasks required by task \(x_i\) (i.e., the tasks belonging to the set \(R_i\)). We estimate the performance of a task with the following equation: 

\begin{equation} \label{eq: performance_equation}
Performance(x_i) = \sigma(b_i, x_i)
\prod_{x_j \in R_i}\sigma(b_j, x_j)
\end{equation}

The performance value for tasks can be part of constraints in the following way. There may be constraints to express that a task \(x_i\) must have a minimum performance \(k\) (where \(k \in [0, 1]\)):
\begin{align*}
C_k: Performance(x_i) \geq k
\end{align*}

The performance value can also be combined with requirement constraints to represent that behavior method \(b_i\) for a task requires a minimum performance of another task \(x_j\) (for example, the execution of a behavior method for motion control may require self-localization with a minimum performance of 0.9). This can be expressed in the following way:
\begin{align*}
C_k:(x_i=b_i)\rightarrow[(x_j \neq \varnothing) \land Performance(x_j) \geq k]
\end{align*}

\section{Constraint satisfaction algorithm} \label{sec: constraint satisfaction algorithm}
The basic strategy followed by our algorithm for behavior coordination is \textit{backtracking search} \cite{vanBeek2006}. This strategy is complete because it guarantees that a solution is found if one exists but, in general, requires exponential amount of time. 

In order to improve search efficiency, we have used additional techniques. For example, we perform \textit{constraint propagation} at each node of the search tree (using the algorithm GAC3 \cite{Mackworth1977}) to remove local inconsistencies and prune the search tree. 

The search efficiency is also improved using mechanisms that reduce the search space (with initialization procedures and variable grouping) and using branching heuristics that are specific of our problem.

In addition to that, behavior coordination is done as a constraint satisfaction \textit{optimization} problem \cite{Tsang1993} to maximize a multi-objective function using the representation about degrees of suitability of behaviors.

Finally, we use two other techniques to (1) coordinate multiple tasks requests with different levels of priority and (2) to manage tasks with reactive start.

The following sections describe in detail these specific techniques.

\subsection{Initialization procedure}

An initialization procedure is used to remove unnecessary values from the domains of variables before starting the search. In practice, this procedure considerably reduces the number of combinations to be explored and, consequently, the time to find a solution.

To formulate the initialization procedure, we use the notation \(domain(x_i) \subseteq D(x_i) \) to represent the current valid domain of \(x_i\) with values that normally change during the search process as constraints are revised. We denote with \(S(x_i)\) the set of behaviors \(b_i \in D(x_i)\) that can be activated in the current situation of the environment (i.e., behaviors that satisfy the assumptions about the environment conditions). The set \(S(x_i)\) is used to avoid considering behaviors that are not compatible with current state of the environment.

The initialization procedure distinguishes two situations: (1) task request, i.e., when behavior coordination is done as an answer to a request to start or stop a task (with a certain priority level) or (2) change in behavior execution, i.e., when behavior coordination is done as an answer to a change in the execution of a behavior (e.g., execution termination due to an unexpected situation).  

In the case of a task request, the domains of variables are initiated in the following way. If the task request asks to start task \(x_i\), the initial domain for variable \(x_i\) is \(domain(x_i) = S(x_i)\). Conversely, if the task request asks to stop task \(x_i\), the initial domain for variable \(x_i\) has a domain that only contains the inactive value \(\varnothing\), i.e., \(domain(x_i) = \{\varnothing\}\). 

In the case of a change in behavior execution, the domains of variables are initiated as follows (here, \(b_i\) identifies the behavior whose execution has changed and \(x_i\) identifies a task performed by this behavior):

\begin{itemize}
\item If behavior \(b_i\) finishes successfully, the initial domain for variable \(x_i\) only contains the inactive value \(\varnothing\), i.e., \(domain(x_i) = \{\varnothing\}\). 
\item If behavior \(b_i\) finishes due to a failure, the initial domain for variable \(x_i\) has a domain that contains all possible values but the behavior that has finished with failure, i.e., \(domain(x_i) = \{\varnothing, b_1, ..., b_{i-1},b_{i+1},..., b_k\}\)  where \(S(x_i) = \{ b_1, b_2, ..., b_k\} \). This initialization prevents that the same behavior can be selected again.
\item If behavior \(b_i\) finishes due to a situation change, the initial domain for \(x_i\) is \(domain(x_i) = \{\varnothing\} \cup S(x_i)\). This initialization allows that the same behavior can be selected again (although with lower performance).
\end{itemize}

In both cases (task request or behavior execution change), the rest of the tasks are initialized as follows:

\begin{itemize}
\item All tasks \(x_j \in X\),  \(j \neq i\) that have the property start on request and are not running, are initialized as \(domain(x_j)=\{\varnothing\}\). The property start on request is useful to remove all candidate behaviors from the initial domain, when the task is not running.
\item The rest of the tasks \(x_j \in X\),  \(j \neq i\) have the initial domain  \(domain(x_j) = \{ \varnothing\} \cup S(x_j)\).
\end{itemize}

\subsection{Grouping variables into independent sets}

The search space is also reduced by separating the set of variables into independent groups. This technique uses the constraint graph, i.e., the graph with variables as vertices and constraints as edges. In general, this graph is disconnected and it can be divided into connected sub-graphs. Variables are grouping into subsets \(X_i \subseteq X\), where each subset includes the variables that belong to a connected sub-graph. This grouping method divides the set of variables X into a partition \(X = X_1 \cup X_2 \cup ... \cup X_r\). This separation can be done in compilation time by analyzing the static constraint graph structure, so that it does not affect the processing time during the search developed for constraint satisfaction.

In our problem, we implement this idea by initializing the domains of variables in a particular way. For example, if the task \(x_i\) is requested to be started, then the domains of the variables that do not belong to the subset of connected variables of \(x_i\) are initiated with a single value that corresponds to the current situation. This initialization technique is also applied when there is a change in behavior execution.

\subsection{Management of priority levels}

As it was mentioned above, each task request has a level of priority. The behavior coordination process has to ensure that requested tasks that are running should not be stopped by other tasks requests that have lower priority level. 

Our algorithm manages the priority levels by filtering the initial domains of variables in the following way. Let consider that the task \(x_i\) is requested to start with priority level \(p\). All tasks \(x_j, j \neq i\) that are running and were requested with higher priority than the priority level \(p\) must remove the value \(\varnothing\) in their initial domain. 

In addition to this technique, we also use another technique to manage priority levels in a different situation. This second technique is applied when a behavior finishes its execution with failure and the behavior corresponds to a task that was not requested (for example, a task that was automatically started because another behavior required its execution). The idea of this technique is to repeat multiple times the constraint satisfaction process, increasing the priority level step by step until a solution is found. 

In more detail, this is done as follows. Initially, we apply the constraint satisfaction process after removing the value \(\varnothing\) in the initial domain of all running tasks that were requested with a level of priority higher than a minimum level \(p\) (e.g., \(p = 0\)). If no solutions are found, we repeat the constraint satisfaction process but, now, the value \(\varnothing\) is removed in the initial domain of the requested tasks with a level of priority higher than \(p+1\). This process is repeated increasing the priority until a solution is found. The effect of this technique is that the solution found may stop some of the requested tasks that are running (due to the behavior failure) but it ensures that these tasks were started with the lowest priority.

\subsection{Branching heuristics}

During the backtracking search process, a tree is developed where each branch corresponds to the assignment of a value to a variable. In our algorithm, the next variable \(x_i \) to branch is determined randomly, but the ordering of the branches of the tree (i.e., the values \(v_i\) that are first assigned to the variable \(x_i\)) is determined by a set of heuristics. 

The following branching heuristics determine which value of the domain of variable \(x_i\), \(domain(x_i)\), is selected first:
\begin{itemize}
\item If task \(x_i\) is not running and \(\varnothing \in domain(x_i)\), then select the value \(\varnothing\). This heuristic gives preference to the current situation, keeping the task not running.
\item If task \(x_i\) is running, its current active behavior satisfies \(b_i \in domain(x_i)\) and the rest of the potential behaviors for task \(x_i\) are less or equally suitable than \(b_i\), then select the value \(b_i\). This heuristic gives  preference to the current situation, selecting the behavior that is performing the task.
\item If task \(x_i\) is running, its current active behavior satisfies \(b_i \notin domain(x_i)\) and there is at least a value in \(domain(x_i)\) corresponding to a behavior, then select first the best suitable behavior. This heuristic tries to keep started the task, but in this case with a different behavior (i.e., the value \(\varnothing\) is not selected in this case).
\item Otherwise, it is selected the behavior with maximum suitability. If two behaviors have the same maximum value, select the behavior randomly.
\end{itemize}

\subsection{Constraint revision}

Our algorithm performs constraint revision according to the general arc-consistency algorithm GAC3 for constraint propagation. However, the revision of performance constraints is a exceptional case that needs a special treatment. This is because the performance value depends on the assigned values for each variable. The algorithm GAC3 does not include constraints with performance values in the constraint queue when the performance changes and, therefore, the general algorithm does not work correctly. The following paragraphs explain how we have done this treatment for this special case. 

According to equation (\ref{eq: performance_equation}), the performance value for each task can be evaluated once a candidate assignment has been found. Our solution postpones the analysis of performance constraints until backtracking search finds a first assignment \(A\). During backtracking search, the performance of each task \(x_i\) is evaluated using the following simplified equations:

\begin{displaymath}
Performance(x_i) = \sigma(b_i^*, x_i)
\end{displaymath}

\begin{displaymath}
b_i^*= 
\argmax_{b_i \in domain(x_i)}
[\sigma(b_i, x_i)]
\end{displaymath}

Once a candidate assignment \(A\) is found, the performance of all tasks is updated using equation (\ref{eq: performance_equation}) and the performance constraints are revised again. If one of these constraints is violated, the assignment \(A\) is rejected and a no-good constraint corresponding to the undesired performance is added to avoid repeating an assignment with these values. This means that, for example, if the violated constraint is \(C_k:(x_i=b_i)\rightarrow[(x_j \neq \varnothing) \land performance(x_j) \geq p]\), then the no-good constraint includes the assignment for the variables of the set \(R_i\) and the assignment for the variable \(x_i\). For instance, if we have the assignment \(A = \{ x_1 = b_1, x_2 =  \varnothing,  x_3 = b_3,  x_4 = b_4, x_5 = b_5\}\), the set of required tasks \(R_1 = \{x_3, x_4\}\) and the violated constraint is \(C_1:(x_1=b_1)\rightarrow[(x_3 \neq \varnothing) \land performance(x_3) \geq 0.8]\), then the no-good constraint is \(no\_good: \{ x_1 = b_1,  x_3 = b_3,  x_4 = b_4\}\).

\subsection{Optimization}

Optimization is performed by repeating backtracking search multiple times to obtain as a final result the assignment that maximizes a multi-objective function. Each time a valid assignment \(A\) is found, backtracking search is repeated but including the next time a no-good constraint with assignment \(A\) to avoid generating the same result. The number of times that backtracking search is repeated is limited by parameters (maximum time or maximum number of solutions) to avoid undesired delays. As a result of this process, multiple assignments \(A\) are generated. The best assignment is the one that maximizes a multi-objective function defined with several partial objective functions that are sequentially maximized.

The first objective function \(f_1(A)\) is the ratio of satisfied requests that activation \textit{A} generates. This objective function gives more value to assignments that satisfy all the tasks requests. In the following equation, \(r\) is the number of satisfied requests and \(n \) is the number of requests done until now that are still active:
\begin{displaymath}
f_1(A) = 
\frac{r}{n}
\end{displaymath}The second objective function \(f_2(A)\) is based on the suitability of the active behaviors:
\begin{displaymath}
f_2(A) = 
\prod_{\forall b_i  active}
\sigma(b_i, x_i)
\end{displaymath}The third objective function is  based on the number of tasks that are running (here we only consider tasks without the property start on request). In the following equation, \(t\) is the total number of tasks and \(a \) is the number of tasks that are active:
\begin{displaymath}
f_3(A) = 
\frac{t - a}{t}
\end{displaymath}The fourth objective function is related to the number of changes that the activation \(A\) generates with respect the current situation. In the following equation, \(c\) is the number of changes:

\begin{displaymath}
f_4(A) = 
\frac{1}{1+c}
\end{displaymath}

\subsection{Delayed start of reactive tasks}

As it was explained above, there may be tasks with the property reactive start. This property means that a task should be started whenever the conditions of the environment make it possible and they are compatible with running tasks.

To manage the way these tasks are started, we use a technique that delays the time when they are started to avoid unnecessary reactive starts between consecutive task requests. For example, the execution of a mission plan may request to start the sequence of tasks \(\langle FollowPath, Rotate\rangle\). If this delay was not used, the reactive task \(Hover\) would be started and stopped between \(FollowPath\) and \(Rotate\).

\begin{figure*}[htb!]
\begin{center}
\includegraphics[width=0.61\textwidth]{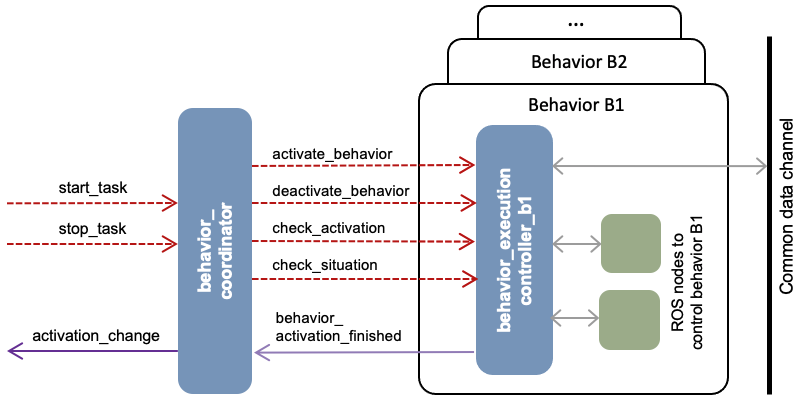}
\caption{ROS-based implementation of \textit{Behavior Coordination CBC}. In this figure, round squares in blue represent the ROS nodes of this software tool, red arrows represent ROS services and purple arrows represent ROS topics.}
\label{fig:ROS-based implementation}
\end{center}
\end{figure*}

This technique works as follows. For each task \(x_i\) (e.g., \(FollowPath\)) we define the set \(T_i\) that includes all tasks that have the reactive start property and are incompatible with \(x_i\) (e.g., the task \(Hover\)). This set is used in to update a queue \(Q\) of tasks in the following way. When a task \(x_i\) finishes its execution, each task \(x_j \in T_i\) is included in the queue \(Q\) with a label that indicates the scheduled time when the task should be started. This time is the current time plus a constant time increase \(\Delta t\) that delays its start. Conversely, when a task \(x_i\) is started, each task \(x_j \in T_i\) is removed from the queue \(Q\) (if it is present in the queue).

The queue \(Q\) is processed at the end of each cycle of behavior coordination. The behavior coordination process operates continuously, generating changes in the set of active behaviors as an answer to the events (about task requests or behavior execution changes). In each cycle, the list of events that have arrived in the last time interval are processed sequentially. Then, the queue \(Q\) is processed to start the reactive tasks whose scheduled time has expired. These tasks are started with a minimum priority level (\(p = 0\))

\section{ROS-based software tool } \label{sec: ros based implementation}

The algorithm described in this paper has been implemented as a software tool called \textit{Behavior Coordinator CBC} that is open source and available\footnote{\url{https://github.com/cvar-upm/coordinator_cbc}} for the general public. This section describes how this tool has been programmed using ROS.

Figure \ref{fig:ROS-based implementation} shows how the functions for self adaptation are programmed using ROS nodes. There is a node called \verb|behavior_coordinator| that implements the constraint satisfaction algorithm presented in this paper. This node interacts with a set of nodes called behavior execution managers. Each behavior execution manager encapsulates the functions related to execution monitoring and activation management for each robot behavior.

\subsection{Behavior execution manager}

The behavior execution manager provides a uniform interface for behaviors to be activated or deactivated, and to inform about their successful or failed termination. 

Each behavior execution manager interacts with other ROS nodes using inter-process communication mechanisms used in ROS. In particular, it provides the following request-reply ROS services:

\begin{itemize}
\item \verb|activate_behavior|: This service activates the execution of the behavior. 
\item \verb|deactivate_behavior|: This service terminates the behavior execution.
\item \verb|check_activation|. This service checks whether the behavior is active or not.
\item \verb|check_situation|. This service verifies that the behavior can be activated in the current situation of the environment (for example, to activate the behavior take off, an aerial robot must be landed).
\end{itemize}

The behavior execution manager publishes a message in the ROS topic called \verb|behavior_activation_finished|. This message is sent when the execution of the behavior has
finished and contains the behavior identification and the cause
of termination. The possible causes of termination that are considered in our method are: goal achieved, time out, wrong progress, situation change, process failure or interrupted. More details about  behavior execution managers can be found in the published work \cite{Molina2021}, which also describes a software library\footnote{\url{https://github.com/cvar-upm/behaviorlib}} to program this type of node.

\subsection{Behavior coordinator}

The behavior coordinator node interacts with the rest of nodes in the following way. It provides two request-reply ROS services called  \verb|start_task| and \verb|stop_task| to accept tasks requests. In addition this node is subscribed to the ROS topic \verb|behavior_activation_finished| to receive  messages that notify when and why a behavior has finished their execution. The behavior coordinator uses these communication mechanisms to receive the events (task requests and behavior execution changes) that execute the constraint satisfaction algorithm to generate new configurations of active behaviors.

The behavior coordinator activates and deactivates behaviors using ROS services provided by behavior execution managers presented above. In addition, the behavior coordinator publishes messages in the ROS topic \verb|activation_change| to inform about changes produced in the set of active behaviors.

\begin{figure}[htb!]
\begin{center}
\includegraphics[width=0.35\textwidth]{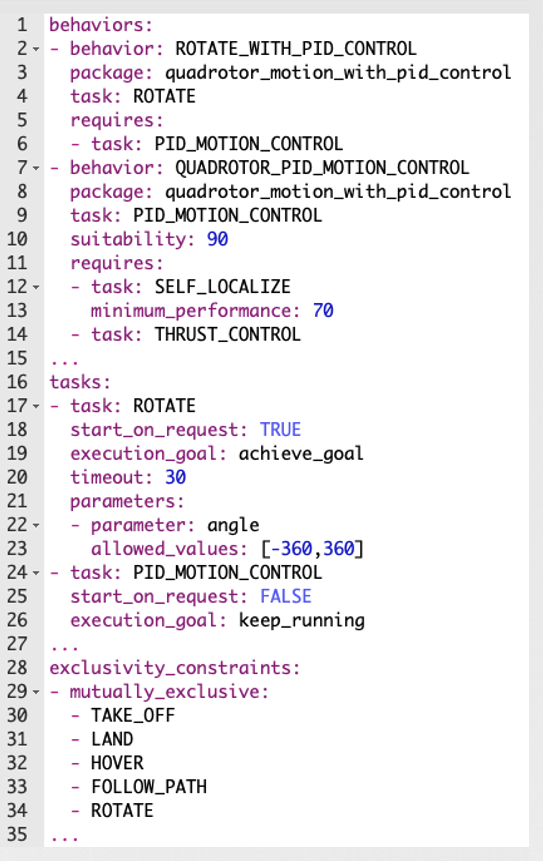}
\caption{Partial example of a behavior catalog written in YAML.}
\label{fig:catalog}
\end{center}
\end{figure}

The behavior coordinator uses as input data a text file called \verb|behavior_coordinator_catalog.yaml| that includes the constraint-based knowledge representation for behavior coordination described in this paper. Figure \ref{fig:catalog} shows a partial example of this catalog (this example corresponds to a simplified implementation in which a behavior can be associated to one single task instead to multiple tasks). The catalog is written manually by the developers of each robotic system considering the specific behaviors used in the robotic application. The file specifies the available behaviors together with the set of compatibility and requirement constraints. YAML syntax is used to make it easier for developers to write the catalog content.

The implementation of the behavior coordinator includes also a procedure that monitors the timeout of behavior execution (which is part of behavior execution monitoring). This procedure is useful to increase robustness because it is able to detect a type of fault that cannot be detected by behavior execution managers (e.g., when a the behavior execution manager node is dead due to a software failure).

\subsection{Behavior execution viewer}

\begin{figure}[htb!]
\begin{center}
\includegraphics[width=0.45\textwidth]{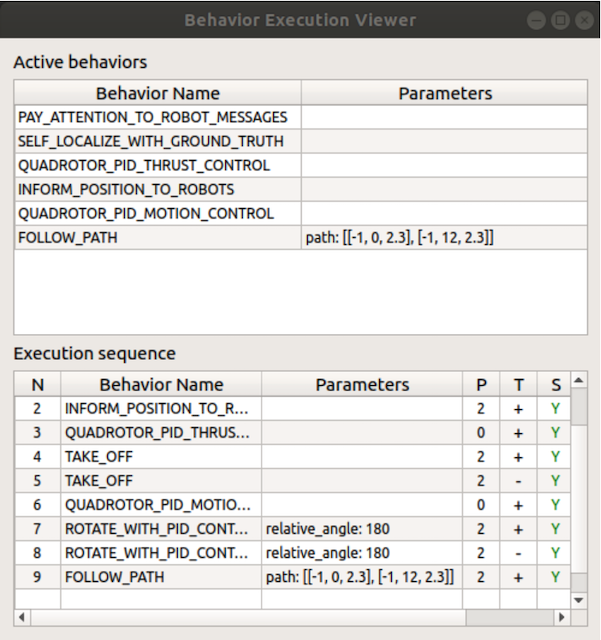}
\caption{Example window presented by the behavior execution viewer.}
\label{fig:behavior execution viewer}
\end{center}
\end{figure}

In addition to the previous ROS nodes, the software tool provides the ROS node called \verb|behavior_execution_viewer| to help the operator monitor how the behavior coordinator activates robot behaviors. Figure \ref{fig:behavior execution viewer} shows an example of window presented by this viewer. The viewer shows at the top the list of active behaviors. At the bottom, the viewer shows the sequence of activations and deactivations with three columns in the right hand side with the following meaning: column P describes the priority level, column T indicates activation (+) or deactivation (-), and column S describes success (Y) or failure (N).

\section{Evaluation} 

\begin{figure*}[htb!]
\begin{center}
\includegraphics[width=0.68\textwidth]{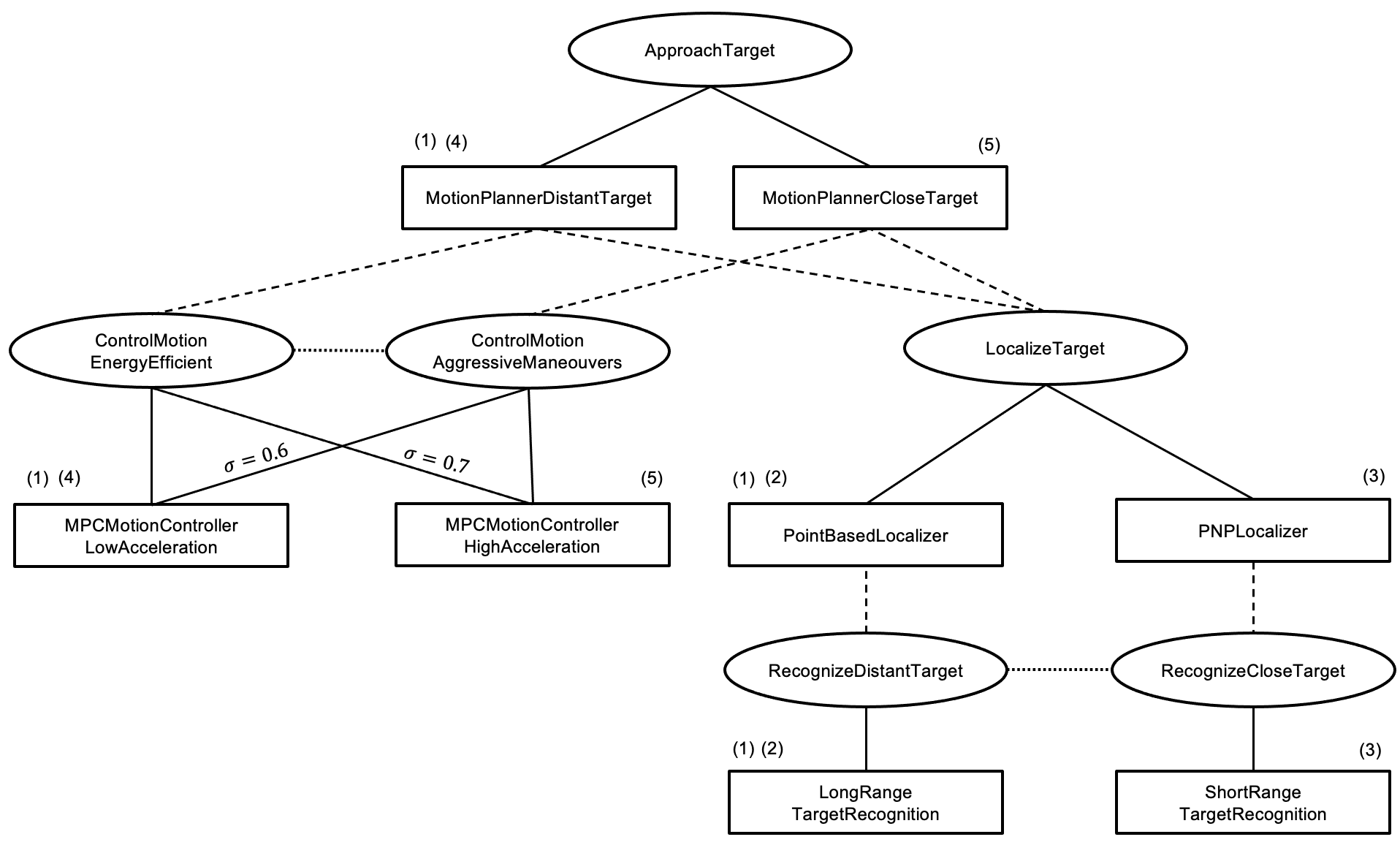}
\caption{Representation for behavior coordination used for the vision-based target following problem. In this figure, ellipses represent behavior tasks and rectangles represent behavior methods. Continue lines represent the relation performed-by, dashed lines represent requirement constraints and dotted lines represent compatibility constraints. For example, the behavior method \textit{PNPLocalizer} requires the task \textit{RecognizeCloseTarget}. Numbers in parentheses represent the order in which they are activated and deactivated.}
\label{fig:task decomposition mbzirc}
\end{center}
\end{figure*}

This section describes the evaluation of the solution presented in this paper. The evaluation covers three aspects. First, we analyze the applicability of the proposed solution to a robotic application that requires self-adaptation. This application corresponds to the  autonomous robot that performs a vision-based target following mission in the competition MBZIRC. Second, we analyze the generality and reusability of the presented algorithm for multiple robotic applications. In this case, we discuss our experience in using this algorithm for different applications in aerial robotics. Finally, we evaluate the performance efficiency of our algorithm in terms of processing time and resources utilization (memory consumption and CPU usage).

\subsection{Applicability to a target following problem}

In order to analyze the applicability of the solution presented in this paper to self-adaptive robots, this section uses the example mentioned at the beginning of this paper that corresponds to an aerial robot developed in our research group \cite{Suarez2020} to participate in the Challenge 1 of the International Competition MBZIRC 2020. In this challenge, an autonomous robot has to catch in less than 15 minutes a ball carried by an aerial vehicle that moves at a speed between 5m/s and 8 m/s. Figure \ref{fig:drones mbzirc} shows (a) the target aerial vehicle with the ball, and (b) the autonomous robot built by our team to perform this task. 

As it was explained in the introduction of this paper, this system exhibits self-adaptive behavior in two aspects. First, when the robot is flying far away from the target vehicle, it moves with slow accelerations to save energy. However, when the robot is close to the target, it performs aggressive maneuvers with high speeds and accelerations in order to be able to intercept the target. Second, this robot uses two different visual detection algorithms to recognize the target vehicle (calibrated to different distances to the target). Since both detection algorithms cannot be executed at the same time due to their high computational load, the robot changes the visual recognition method according to the distance to the target.

Figure \ref{fig:task decomposition mbzirc} shows a model for behavior coordination for this system, using the representation described in this paper. In this model, the task {\footnotesize\verb|ApproachTarget|} can be done by two different behaviors depending on the distance to the target. In addition, each one of these behaviors requires other tasks. For example, the behavior {\footnotesize\verb|MotionPlannerCloseTarget|} requires two tasks: {\footnotesize\verb|ControlMotionAggressiveManeuvers|} and {\footnotesize\verb|LocalizeTarget|}. In its turn, the task {\footnotesize\verb|ControlMotionAggressiveManeuvers|} can be performed by two behaviors but one of them ({\footnotesize\verb|MPCMotionControllerHighAcceleration|)} is more suitable for this task (with the measure 1.0) than the other method, which has a measure of 0.6.

To illustrate how this model is used for self-adaptation, let suppose that behaviors with number (1) are active in a particular moment during the mission execution, performing the task {\footnotesize\verb|ApproachTarget|} (in this moment, the target vehicle is distant). As the mission execution progresses, the autonomous robot gets closer to the target. When the target is close enough, the situation conditions for behavior {\footnotesize\verb|LongRangeTargetRecogniton|} are not satisfied (e.g., these conditions establish the assumption that the target image should be small) and this behavior terminates its execution. Then, as a response to this event, the behavior coordination process looks for an alternative configuration of active behaviors to localize the target. The new configuration is obtained by deactivating behaviors with number (2) and activating behaviors (3). As a consequence of this change, the situation conditions for behavior {\footnotesize\verb|MotionPlannerDistantTarget|} are not satisfied (the situation conditions use the spatial coordinates estimated by the target localization method) and this behavior terminates its execution. As a response to this event, the behavior coordination process generates another configuration which deactivates behaviors (4) and activates behaviors (5). As a result of this process, the global task {\footnotesize\verb|ApproachTarget|} is finally supported by the active behaviors (3) and (5).

This description shows that the solution presented in this paper for self-adaptation is applicable to the target following problem. We present in this section a second version for self-adaptation that is different from the first version implemented for the competition. The application of the algorithm presented in this paper provides certain advantages. In this second version, the procedure for dynamic selection of alternative methods (the adaptation logic) is not intertwined with the robot behaviors and, as a result, the computer programs that implement control algorithms are simpler and more reusable for other applications. In addition, the mission specification is simpler than the specification in the first version because it does not include decisions related to the dynamic selection of methods.

\begin{figure*}[htb!]
\begin{center}
\includegraphics[width=0.79\textwidth]{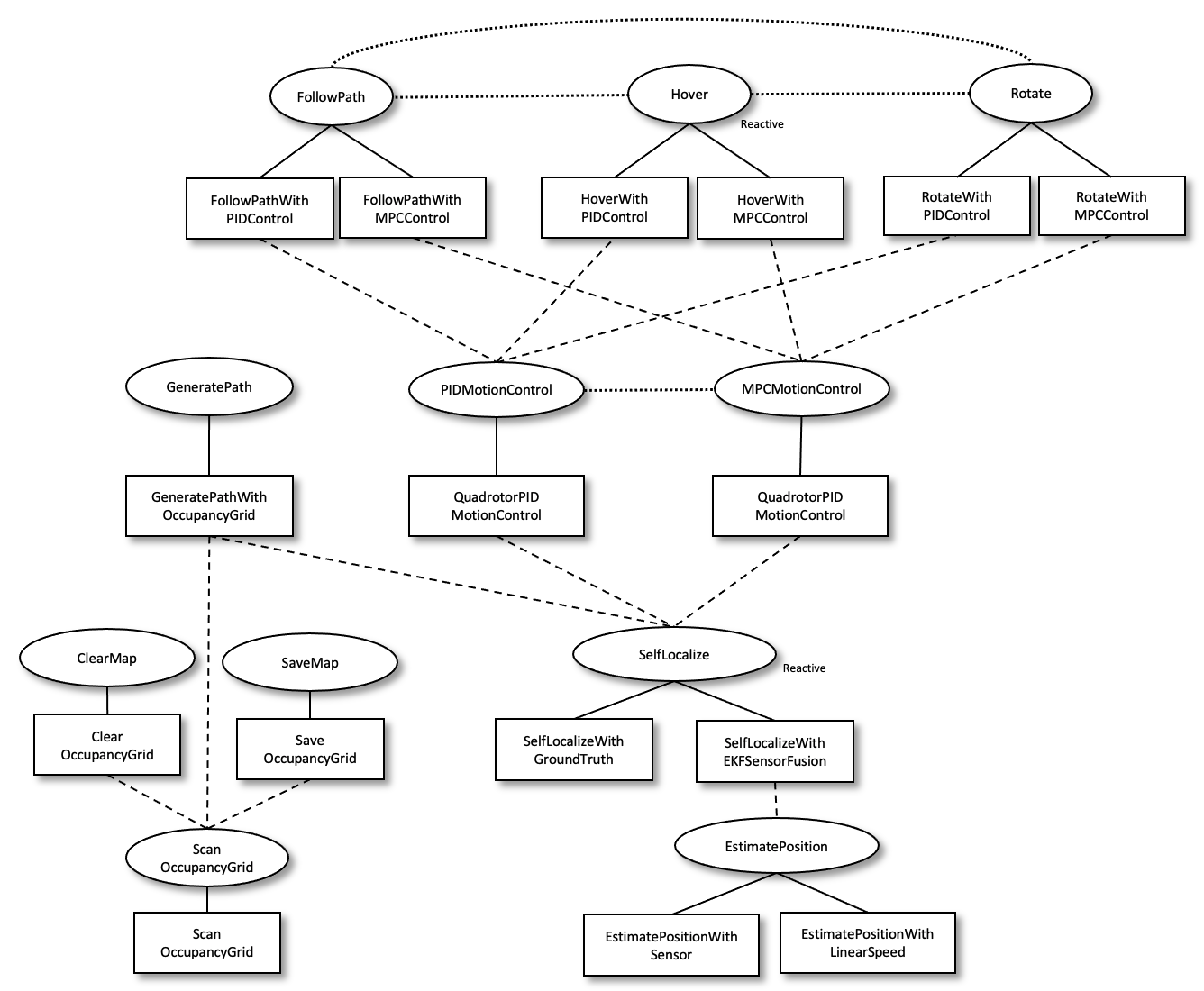}
\caption{Partial view of the model for behavior coordination used for building aerial robotic applications. The same model was reused for the development of thirteen demonstrative applications.}
\label{fig:example network catalog}
\end{center}
\end{figure*}

\subsection{Reusability for multiple aerial robotic applications}

The solution presented in this paper has been used for building demonstrative aerial robot applications that are part of the Aerostack\footnote{\url{http://www.aerostack.org}} software environment \cite{Sanchez2017}. Version 4.0 of this environment was released with 13 demonstrative applications of aerial robots that used the algorithm described in this paper. For instance, Figure \ref{fig:example aerostack project} shows an image  generated by the execution of an application for airplane inspection (using the Gazebo simulation tool). The image shows in the right hand side (upper part) the window presented by the behavior execution viewer with the current state of the active behaviors and the sequence of behavior activations. Besides this application, Aerostack includes other applications related to exploration of a tower building, warehouse inventory and bridge inspection with multiple aerial robots.  

The experience in the development of such applications demonstrated the generality of our algorithm to be used in multiple robotic systems with different platforms. The algorithm was able to operate in real time with the efficiency required by aerial robots. This experience also showed that the constraint-based representation was easy to write (in the form of a behavior catalog) and the same representation was shared by multiple applications. Figure \ref{fig:example network catalog} shows graphically part of the elements used in this model with multiple tasks and behaviors. 

In general, unmanned aerial vehicles are a type of mobile robots that may require self-adaptation in multiple situations to support effective autonomous behavior. For instance, an aerial vehicle can use different aerial motion controllers that can be dynamically selected according to the variable weight (e.g., a cargo drones or aerial robotic platforms with grasping mechanisms \cite{Hingston2020}), or according to the expected response time of  the task to be done. An aerial robot can also use different self-localization methods (e.g., GPS localization or alternative visual-based methods using different types of cameras) and they can be dynamically selected according to the particular characteristics of the environment (outdoors or indoors, distance to obstacles, etc.).

\subsection{Performance efficiency evaluation}

This section describes results of experiments conducted to evaluate the performance efficiency (e.g., processing time, CPU usage and memory consumption) of the algorithm presented in this paper for behavior coordination. The characteristics of the computer used for the experiments are the following: CPU AMD Ryzen 7 3800x, 8 cores, 16 threads, 3.9 GHz and 16 GB RAM 3200 MHz.

\begin{figure*}[htb!]
\begin{center}
\includegraphics[width=0.67\textwidth]{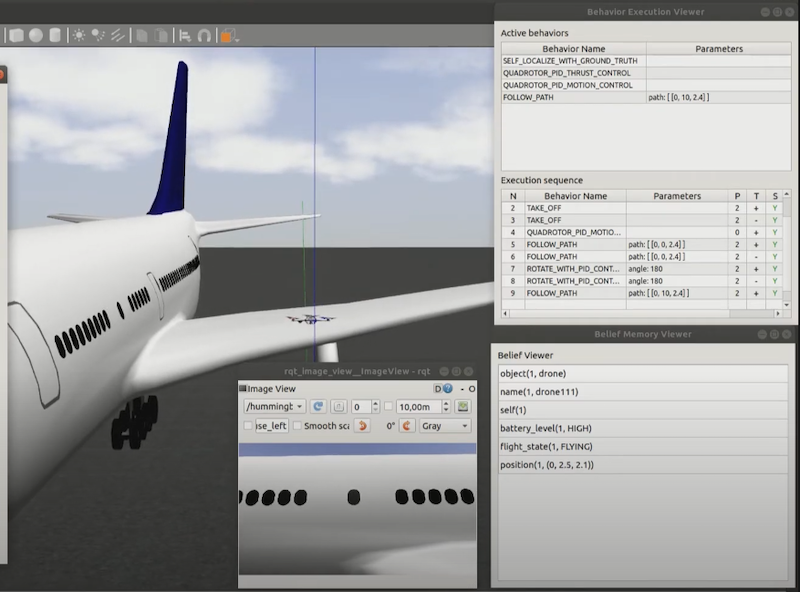}
\caption{Screenshot of a demonstrative application in aerial robotics for autonomous inspection of the surface of an airplane. This application uses Gazebo to simulate the aerial robot and the environment.}
\label{fig:example aerostack project}
\end{center}
\end{figure*}

Table \ref{tab:processing time applications} shows the processing time corresponding to two demonstrative applications of Aerostack. The first one (application 1) is the application for autonomous inspection of airplane surfaces mentioned above (Figure \ref{fig:example aerostack project}). The second application (application 2) corresponds to a problem where an aerial robot explores an unknown building. In this second application, the robot uses a path planner and a lidar sensor to detect obstacles. The path planner updates the path to be followed when new obstacles are detected. In both applications the robot and the environment are simulated using Gazebo. The columns of table \ref{tab:processing time applications} are \(m\) (maximum number of solutions for behavior coordination), \(n\) (number of times that behavior coordination was executed during the mission execution), and the processing time \(t\) taken by behavior coordination in milliseconds (mean, minimum and maximum values). The results obtained in these experiments show that the processing time of the algorithm is sufficiently low to be used in this type of missions.

Table \ref{tab:processing time search space} shows the processing time considering different sizes of the search space. The first row of this table shows the values corresponding to the behavior catalog used in Aerostack for multiple aerial robotics applications. In these experiments, we used other three catalogs created artificially to estimate the influence of the size of the search space in the processing time. These catalogs included abstract behaviors and tasks that were related with multiple candidate behaviors for each task, and multiple tasks required by each behavior. 

The table shows the number of constraints of these catalogs (column \(c\)) and the size of the search space (column \(s\)). The processing time is presented in milliseconds considering two cases: time required to find one solution (\(m = 1\)) and time required to find the first five solutions  (\(m = 5\)). The size of the search space is \(s = \prod_i|D(x_i )|\), where \(|D(x_i )|\) is the number of values of the domain of task \(x_i \). For the reference catalog, the size of search space is low (288 combinations) due to the initialization procedure of domains. In the case of the other three catalogs we did not apply any initialization procedure in order to analyze how the processing time increases with high numbers of combinations. 

The results show that our algorithm is able to find efficiently one solution (in less than 1 ms) when the number of combinations is \(s \leq  10^{7}\). This  is an acceptable limit as it is shown by the applications developed with the solution presented in this paper (e.g., the number of combinations is reduced to 288 after initialization) and shows that there is scope for the algorithm to be used in more complex applications. 

The results also show how the processing time increases when the algorithm searches for multiple solutions (\(m = 5)\). For example when \(s = 1.7 \times 10^{7}\) the processing time increases to 3.9 ms. As it was mentioned, this time increase can be handled by using a parameter that defines a processing time limit. The search for multiple solutions stops when the time consumed exceeds this limit.

\begin{table}
\caption{Processing time in aerial robotic applications.}
\label{tab:processing time applications} 
\centering
\footnotesize
\begin{tabular}{ | c |c | c | c |c |c |}
\hline
Application & \(m\) & \(n\) & \(t_{mean}\) & \(t_{min}\) & \(t_{max}\)\\
\hline
1 & 1 & 18 & 0.14 & 0.12 & 0.18 \\ 
\hline
1 & 10 & 18 & 0.22 & 0.16 & 0.31 \\ 
\hline
2 & 1 & 311 & 0.20 & 0.03 & 1.47 \\ 
\hline
2 & 10 & 350 & 0.28 & 0.03 & 2.25 \\ 
\hline
\end{tabular}
\end{table}

\begin{table}
\caption{Processing time in relation to search space.}
\label{tab:processing time search space} 
\centering
\footnotesize
\begin{tabular}{ | c |c | c | c |c |}
\hline
Catalog & \(c\) & \(s\) & \(t \ (m=1)\) & \(t\ (m=5)\) \\
\hline
\hline
Reference & 33 & 288 & 0.2 & 0.6 \\ 
\hline
\hline
1 & 24 & \(1.9 \times 10^4\) & 0.3 & 1.7 \\ 
\hline
2 & 54 & \(1.7 \times 10^7\) & 0.7 & 3.9 \\ 
\hline
3 & 99 & \(6.9 \times 10^{10}\) & 3.5 & 17.8 \\ 
\hline
\end{tabular}
\end{table}

Tables \ref{tab:processing time applications} and \ref{tab:processing time search space} show the time that our algorithm takes to search for solutions, not including the time required to initialize the domains. The initialization time increases linearly with the number of behaviors and it is is mainly affected by the delay generated by the ROS service \verb|check_situation|. In our experiments using application 2, the average initialization time was 21.1 ms.

\begin{figure}[htb!]
\begin{center}
\includegraphics[width=0.45\textwidth]{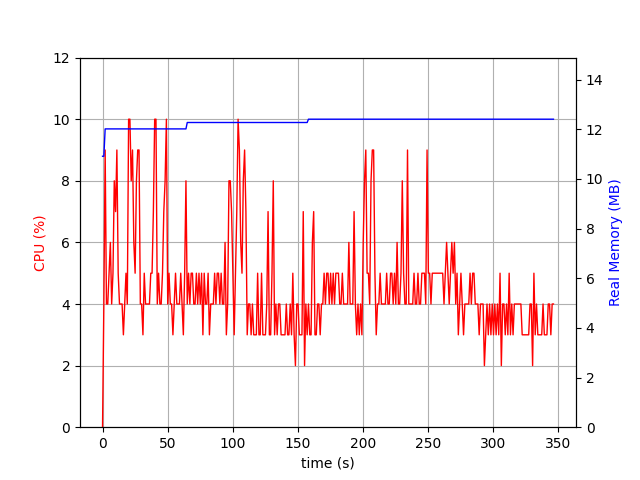}
\caption{Memory consumption and CPU usage of the behavior coordination process.}
\label{fig:resources consumption}
\end{center}
\end{figure}

Finally, Figure \ref{fig:resources consumption} shows the memory consumption and CPU usage corresponding to the ROS node \verb|behavior_coordinator| for application 2 (which demands more resources than than application 1). The values show that this ROS node required 12.4 MB of memory and it used up to 10\% of CPU usage. The value of CPU usage corresponds to one of the 16 threads of the computer used for the experiments.

\section{Conclusion}
This paper has presented an original algorithm to dynamically configure a robot control architecture, which is applicable to the development of self-adaptive autonomous robots. 

In our work, self-adaptation is done by activating and deactivating basic robot behaviors. Our algorithm follows a configuration approach to decide which behaviors should be activated in each situation. 

The algorithm is able to coordinate the execution of multiple behaviors in response to both reactive events (e.g., unexpected events caused by faults) and deliberative events (e.g., planned tasks), considering also multiple sources of tasks requests that are managed with a priority scheme.

The used representation follows a constraint-based configuration approach (e.g., with compatibility and requirement constraints). The algorithm performs backtracking search and constraint propagation (using the general algorithm GAC3) together with other techniques (initialization procedures, variable grouping and branching heuristics) to achieve the performance required by robotic systems. The algorithm also applies an optimization procedure using a representation about degrees of suitability of behaviors. 

The solution has been implemented as a software development tool called Behavior Coordinator CBC (Constraint-Based Configuration), which is based on ROS and open source, available to the general public. 

This tool has been successfully used for building multiple applications of autonomous aerial robots. This practical experience showed that the constraint-based representation was easy to write (in the form of a behavior catalog) and the same representation was shared by multiple applications. In addition, the values about performance efficiency obtained in our experiments showed that the algorithm presented in this paper can be used in applications with more complex models for self-adaptation.





\ifCLASSOPTIONcaptionsoff
  \newpage
\fi

\IEEEtriggeratref{30}


\bibliographystyle{IEEEtran}
\bibliography{IEEEabrv,references}
%







\end{document}